\documentclass{article}

\usepackage{arxiv}

\usepackage[utf8]{inputenc} 
\usepackage[T1]{fontenc}    
\usepackage{hyperref}       
\usepackage{url}            
\usepackage{booktabs}       
\usepackage{amsfonts}       
\usepackage{nicefrac}       
\usepackage{microtype}      
\usepackage{lipsum}
\usepackage{graphicx}
\usepackage[table,xcdraw]{xcolor}

\title{DeepClaw: A Robotic Hardware Benchmarking Platform for Learning Object Manipulation}

\author{
    Fang~Wan \\
        SUSTech Institute of Robotics \\
        Southern University of Science and Technology \\
        Shenzhen, Guangdong 518055, China \\
        \texttt{\protect\href{mailto:sophie.fwan@gmail.com}{sophie.fwan@gmail.com}} \\
    \And
    Haokun~Wang \\
        Department of Mechanical and Energy Engineering \\
        Southern University of Science and Technology \\
        Shenzhen, Guangdong 518055, China \\
        \texttt{\protect\href{mailto:11510135@mail.sustech.edu.cn}{11510135@mail.sustech.edu.cn}} \\
    \And
    Xiaobo~Liu \\
        Department of Mechanical and Energy Engineering \\
        Southern University of Science and Technology \\
        Shenzhen, Guangdong 518055, China \\
        \texttt{\protect\href{mailto:11930807@mail.sustech.edu.cn}{11930807@mail.sustech.edu.cn}} \\
    \And
    Linhan~Yang \\
        Department of Mechanical and Energy Engineering \\
        Southern University of Science and Technology \\
        Shenzhen, Guangdong 518055, China \\
        \texttt{\protect\href{mailto:11950013@mail.sustech.edu.cn}{11950013@mail.sustech.edu.cn}} \\
    \And
    Chaoyang Song \thanks{Corresponding Author.} \\
        Department of Mechanical and Energy Engineering \\
        Southern University of Science and Technology \\
        Shenzhen, Guangdong 518055, China \\
        \texttt{\protect\href{mailto:songcy@ieee.org}{songcy@ieee.org}} \\
}

\begin{document}
\maketitle

\begin{abstract}
We present DeepClaw as a reconfigurable benchmark of robotic hardware and task hierarchy for robot learning. The DeepClaw benchmark aims at a mechatronics perspective of the robot learning problem, which features a minimum design of robot cell that can be easily reconfigured to host robot hardware from various vendors, including manipulators, grippers, cameras, desks, and objects, aiming at a streamlined collection of physical manipulation data and evaluation of the learned skills for hardware benchmarking. We provide a detailed design of the robot cell with readily available parts to build the experiment environment that can host a wide range of robotic hardware commonly adopted for robot learning. We also propose a hierarchical pipeline of software integration, including localization, recognition, grasp planning, and motion planning, to streamline learning-based robot control, data collection, and experiment validation towards shareability and reproducibility. We present benchmarking results of the DeepClaw system for a baseline Tic-Tac-Toe task, a bin-clearing task, and an jigsaw puzzle task using three sets of standard robotic hardware. Our results show that tasks defined in DeepClaw can be easily reproduced on three robot cells. Under the same task setup, the differences in robotic hardware used will present a non-negligible impact on the performance metrics of robot learning. All design layouts and codes are hosted on Github for open access.
\end{abstract}

\keywords{Benchmark \and Manipulation \and Robot Learning}

\section{Introduction}
\label{sec:Introduction}
Robot Learning is an active field of research that adopts a data-driven approach by applying machine learning algorithms to advanced robots for autonomous control towards intelligent and physical interactions. In order to build up the database for model training, researchers usually adopt passive perception such as RGB or depth cameras to formulate an image-based robot control scheme as experiment setup \cite{Levine2016Learning, mahler2017dex, mandlekar2018roboturk, james2019rlbench}. Such an image-based method provides a continuous collection and validation of the interaction data in manipulation tasks. However, active perception may be necessary to capture the localized interaction at the physical interface between the gripper and objects, as well as the functional performance of the multi-joint manipulators.

Unlike robots such as autonomous cars or drones with significant interactions with the road or the air, robotic manipulation involves a wide range of objects with vastly differentiated physical properties and functional behaviors, making it a challenging task to build a learning system for robotic manipulation \cite{OpenAI2019Solving}. With the availability of a wide selection of robotic hardware with differentiated engineering specifications and functional performances, how to make an informed selection and integration of robot hardware becomes the first question to ask when transferring learned models to reality. 

\begin{figure}[htbp]
    \centering
    \includegraphics[width=1\columnwidth]{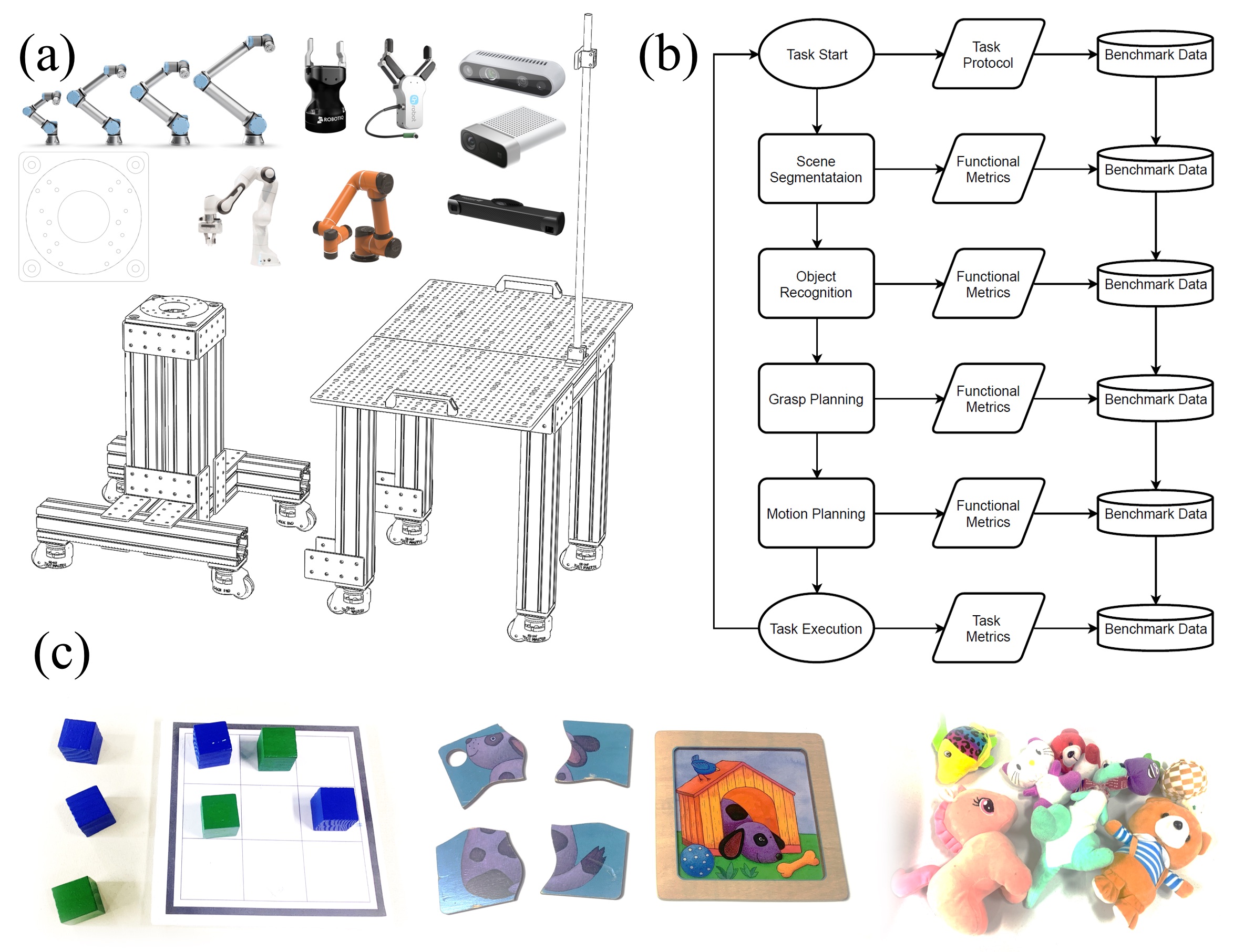}
    \caption{The design of DeepClaw platform towards a reproducible and shareable benchmark for learning robotic manipulation, including (a) a standardized robot station design, (b) a modular pipeline for functional integration, and (c) exemplified task variations.}
    \label{fig:PaperOverview}
\end{figure}

In this paper, we propose the DeepClaw in Fig. \ref{fig:PaperOverview} as a minimum robot cell design to benchmark the robotic hardware for learning object manipulation, which involves three levels of embodiment. 
\begin{itemize}
    \item Modular frame design: aims at a standard mounting frame for all hardware components involved in a robot learning task, which can be easily sourced from global suppliers and local manufacturing shops with open access engineering drawings and assembly instructions for a baseline experiment environment. 
    \item Functional pipeline hierarchy: enables a unified software architecture of system integration, algorithm pipeline implementation, data communication. 
    \item Game tasks \& objects: demonstrates three benchmarking tasks, namely tic-tac-toe, claw machine and jigsaw puzzle, with spatial-temporal task metrics and specific functional metrics for hardware benchmarking.
\end{itemize}

In the rest of the paper, we first review the related work on benchmarks of robot learning in Section \ref{sec:RelatedWork}. In section \ref{sec:DeepClawDesign}, we present a detailed description of the DeepClaw. Section \ref{sec:ExpBenchmark} presents the experiment design and results of three game-based benchmarks using DeepClaw with three sets of robotic hardware. Final remarks are enclosed in section \ref{sec:Conclusion}, which ends this paper.

\section{Related Work}
\label{sec:RelatedWork}
While it remains a question of using machine learning to aid the design of a manipulation system for robot learning \cite{Gealy2019Quasi}, there is a wide range of robotic hardware to build up the initial understanding. Collaborative robots, as a light-weight version of the industrial robots, provide ease of access to robotic manipulation with force-limited safety, cost-effectiveness, and user-friendly interfaces \cite{haddadin2008collision}. Although recent research by Google shows the potential of using large-scale picking data to train models of robot learning \cite{Levine2016Learning}, it is unrealistic to custom build and deploy such a large number of robotic hardware for continuous research. The recent proposal of the RoboNet demonstrates an improved effort using data collected from various robot cells with similar setups and experiment environment in an open-access format \cite{dasari2019robonet}, which focus mainly on visual foresight tasks. Other efforts such as the ROBEL \cite{Ahn2019ROBEL} and REPLAB \cite{Yang2019REPLAB} aim at a low-cost infrastructure for reinforcement learning using robotic hardware with limited motor function. The RLBench is another effort to build a simulated environment for reinforcement learning using a single robot of Franka Emika through 100 task variations \cite{james2019rlbench}.

The broad selection of robotic hardware and a lack of consideration of their mechanical differences motivate the need for a benchmark for robotic hardware for learning object manipulation. A recent survey shows that there are more than 30 collaborative robots with different functional performance and engineering specifications \cite{Robotiq2019Cobots}. To further reduce the dimensionality of the learning problem, parallel grippers \cite{pinto2016supersizing, mahler2017dex}, as a mature, robust, and cost-effective design of end-effector, are widely adopted instead of using robotic hand with multi-fingers \cite{Andrychowicz2019Learning, OpenAI2019Solving}. The object variations and task specifications add further complexity in setting up the robot cell for learning \cite{james2019rlbench}. While many machine learning research much relies on simulated data to test algorithm as a first attempt \cite{memmesheimer2019simitate}, researchers also become increasingly aware of the gap between simulation and reality when transferring the learned model for hardware validations \cite{rusu2016sim, james2017transferring}. 

After reviewing the efforts of making shareable benchmarks for manipulation from literature in Table \ref{tab:BenchmarkReview}, we summarize five aspects of consideration to develop an in-depth understanding of the physical interaction in object manipulation.

\subsection{Manipulation Task}
Manipulation task focuses on a goal-oriented benchmarking of the integration between advanced robotics and machine learning. For arm-type manipulators, the general task design spans from basic human skills such as picking, placing, pushing \cite{dasari2019robonet}, etc., to a wide range of advanced tasks in daily life activities \cite{kemp2007challenges}. A hierarchical decomposition of the manipulation task is usually a preferred choice to establish reusable skills for learning system design \cite{Bohg2014Data}, which requires a systematic task protocol design.

\subsection{Target Object}
The target object is a minimum collection of the physical features in manipulation, including shape, weight, size, color, material, etc., which must be perceived by the robot system with accuracy. It represents an object-centric generalization in robot learning, such as the YCB object dataset \cite{calli2015benchmarking}. Simulated object generation also provides an alternative source of data for offline training \cite{mahler2017dex, mandlekar2018roboturk}.

\subsection{Operating Environment}
The operating environment requires reliable and robust support to maintain a quality data collection protocol and a safe operation of the robot with human, if necessary. It usually involves a flat desk \cite{mandlekar2018roboturk} or an object tray \cite{Levine2016Learning} as the manipulation arena. The robot can be mounted on the desk or a dedicated pedestal with a fixed spatial relationship with the desk \cite{james2019rlbench}. In the meanwhile, one should also consider the mounting frame and angle of the cameras for calibration and view range, which can be on the hand, to the base, or on the robot. 

\subsection{Robotic Hardware}
Robotic hardware usually involves a manipulator as the arm, a vision sensor as the eye, and an end-effector as the hand for an image-based robot control system \cite{dasari2019robonet}. As shown in Table \ref{tab:BenchmarkReview}, while there is a wide range of hardware available for integration, researchers gradually converge to a selection guideline based on collaborative robots, RGB-D cameras, and 2-finger parallel grippers or a single suction cup to build the robot cell. 

\subsection{Algorithm Pipeline}
The algorithm pipeline represents the control logic of the learning system, which relies on a quality collection of the interaction data of the robotic hardware against the target object in the operating environment for a specific manipulation task. The main input data is collected through passive perception such as 2D, 3D, or depth cameras towards an image-based robot control for large-scale data collection. A potential drawback is the limited use of robotic hardware in manipulation tasks and learning system design. 

\begin{table}[htbp]
    \centering
    \caption{Review of shareable manipulation benchmarks with physical and simulated robots (shaded) sorted by year.}
    \includegraphics[width=1\columnwidth]{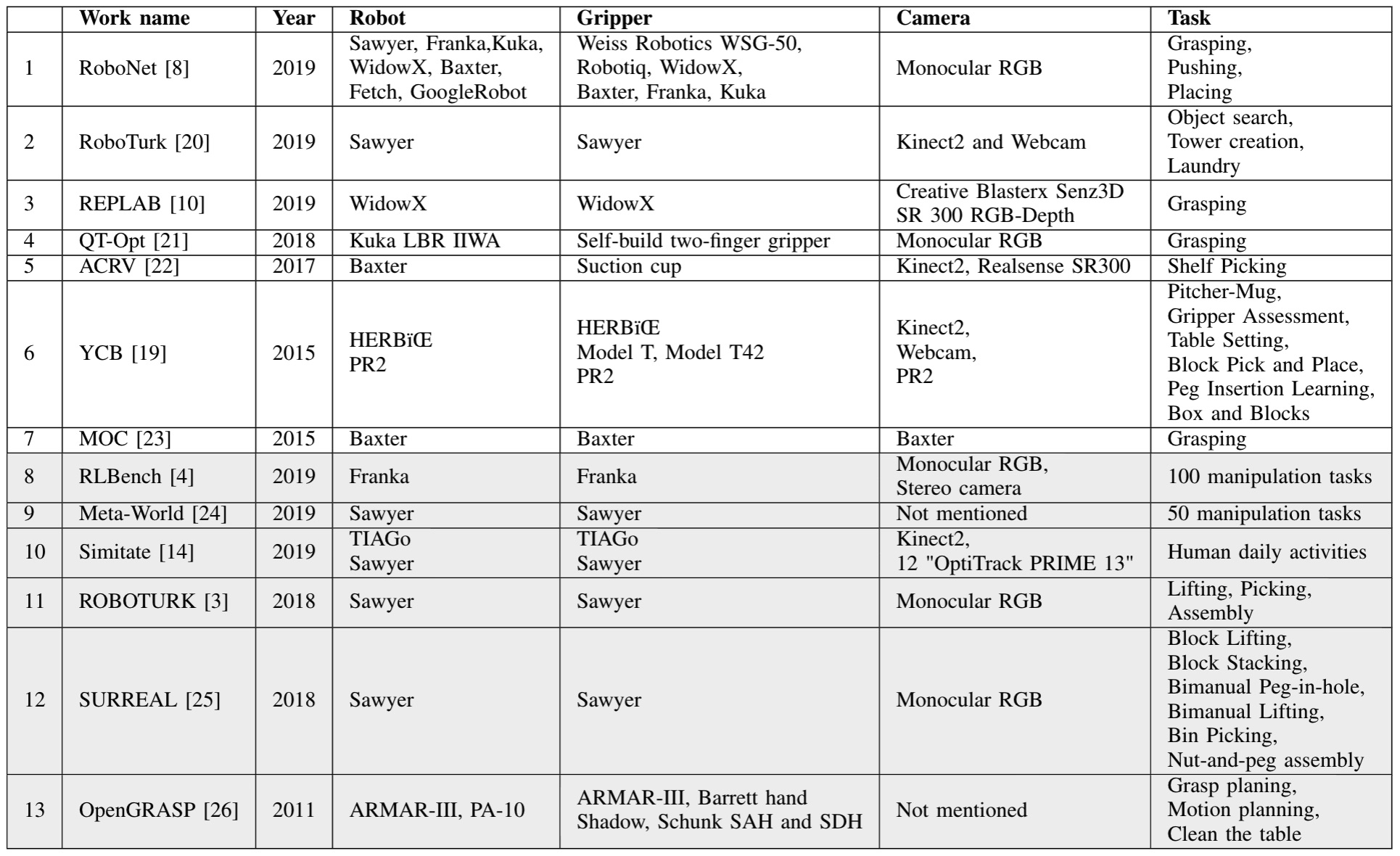}
    \label{tab:BenchmarkReview}
\end{table}

\section{DeepClaw Benchmark Design}
\label{sec:DeepClawDesign}

\subsection{Design Overview}
DeepClaw is a benchmarking system for robot manipulation designed to be modularized, extendable, and easy to use on real robots and the environment. As shown in Fig. \ref{fig:PaperOverview}, DeepClaw consists of four components:
\begin{enumerate}
  \item A standardized robot cell design.
  \item A set of unified driver interfaces that serves as a connection layer between the benchmarking algorithms and the hardware, including robot arms, robot hands, visual sensors, tactile sensors.
  \item A collection of baseline algorithms for segmentation, recognition, grasp planning, and motion planning.
  \item A pipeline for task definition and functional integration.
\end{enumerate}

The standard hardware design and functional pipeline of DeepClaw aim to accelerate the benchmarking process. Tasks defined in DeepClaw will be easily reproduced on different robot hardware setups by changing the configuration files while keeping the same task pipeline. The software is mainly written in Python for ease of use. However, as some hardware suppliers only provide APIs in C++, some of the DeepClaw drivers use a mix of Python and C++ as well. The key features of the design of the DeepClaw are described in the following four aspects. More comprehensive documentation and codes are provided at https://github.com/bionicdl-sustech/DeepClawBenchmark.

\subsection{System Integration}

\begin{figure*}[htbp]
    \centering
    \includegraphics[width=1\columnwidth]{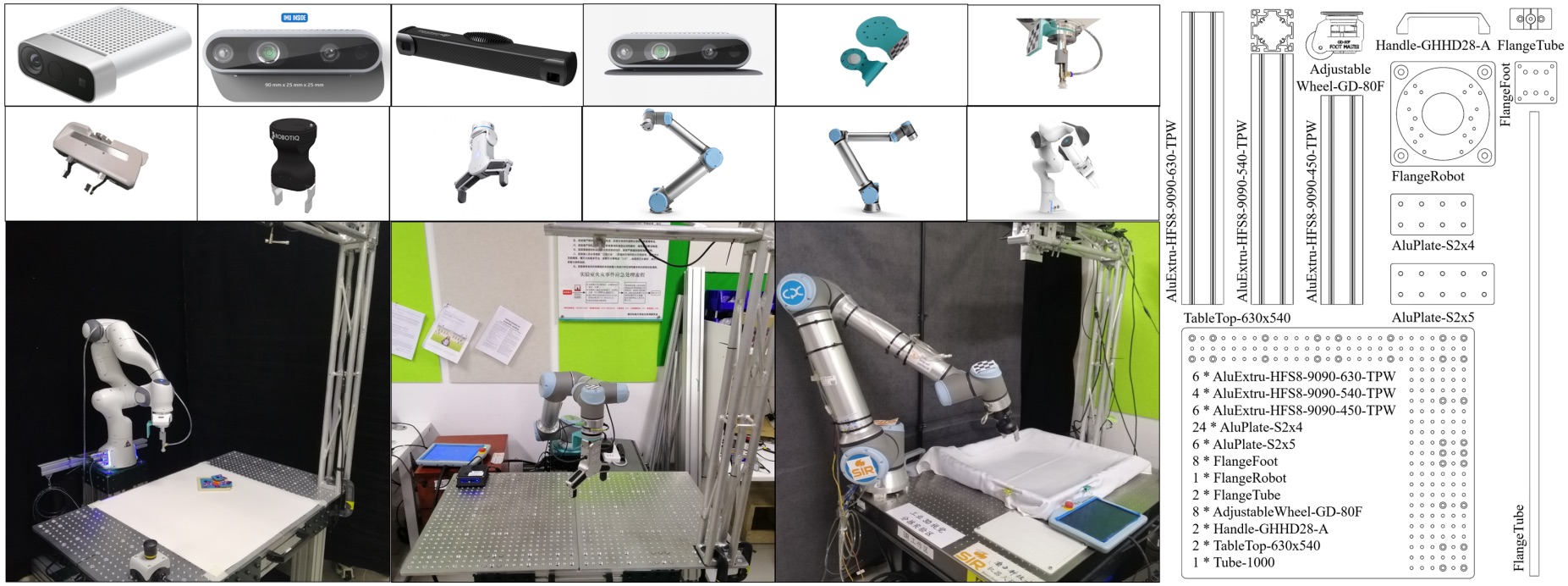}
    \caption{Design overview of the robotic hardware in DeepClaw, including the supported cameras, end-effectors, and robotic manipulators (top), the three robot cells used in this papers (bottom), and the all the mechanical parts for a full assembly (right).}
    \label{fig:SystemIntegration-ExampleHardware}
\end{figure*}

The mechanical design of the DeepClaw station involves aluminum extrusions commonly available from global suppliers such as MISUMI, with part numbers shown on the right of Fig. \ref{fig:SystemIntegration-ExampleHardware}. For the convenience of assembly, we also introduced a few aluminum plates with through holes, which can be easily machined from local shops. A robot mounting plate is also included with threaded holes matching the base mount of the robots used in this paper, including Universal Robots and Franka Emika, which is also suitable for other manipulators from suppliers such as AUBO. We included detailed design files of the DeepClaw on the project Github page so that others can make informed modifications to suit their robot system.

We have implemented quite a few drivers for hardware shown in Fig \ref{fig:SystemIntegration-ExampleHardware} including robot arm drivers for UR5, UR10e, and Franka; robot hand drivers for HandE from Robotiq, RG6 from Onrobot, gripper from Franka, and standard suction cup; visual sensor drivers for Realsense 435/435i, Azure Kinect, PhoXi M from Photoneo; force-torque sensor driver for OptoForce from Onrobot. Based on these hardware devices, we have built three robot stations, as shown in Fig. \ref{fig:SystemIntegration-ExampleHardware}. Robot cell one consists of a assembled station with table top, a Franka robot arm with a suction cup, and a Realsense 435i. Robot cell two consists of a assembled station with table top, a UR5, a RG6 gripper, and a Realsense 435. Robot cell three consists of a manufactured optical table, a Realsense 435, an Azure Kinect, and a Robotiq HandE. All the cameras are mounted about one meter above the table top and facing downward.

The DeepClaw cell, shown on the bottom of Fig \ref{fig:SystemIntegration-ExampleHardware}, is a self-contained robot cell for manipulation tasks, including an assembled station, a robot arm, a robot end-effector, a visual sensor. Thanks to the standardized design, we are able to buy the components and assemble the robot station quickly. The hardware setup is defined in the software part of DeepClaw through configuration files. The same task should be easily reproduced on different hardware setups if the configuration file is adequately defined.

We also design a 6cm$\times$6cm 3D printed calibration board mounting on the tool flange without affecting the end-effectors, as shown in Fig. \ref{fig:SystemIntegration-ExampleHardware}. A 3x3 checkerboard pattern is printed and stick on the calibration board. The hand-eye calibration is implemented by registering the coordinates of the checkerboard for the robot arm and camera. Since the transformation from the tool center point (TCP) to the checkerboard is known, the hand-eye matrix can be easily obtained from the singular value decomposition (SVD) method \cite{Umeyama1991}. Scripts that automatically run the hand-eye calibration and to verify the accuracy of the calibration are provided in the calibration module of the DeepClaw.

\subsection{Tasks vs. Objects}
In DeepClaw, a manipulation task involves three hierarchical concepts: task, sub-task, functionality module. A task protocol should clearly define the main purpose of the task, the target objects, the robot, and hardware setup, procedures to fulfill the task and execution constraints \cite{calli2015benchmarking}. Each task may consist of several repetitive sub-tasks. A pipeline of functional modules can accomplish each sub-task.

The most similarity between game and dexterous manipulations enable reproducible experiments in various environment. All of the game manipulation tasks can be classified from two different perspectives: spatial reasoning and temporal reasoning. Compared with human daily dexterous manipulations, game manipulations have a noticeable distinction in spatial and temporal dimensions. "Jigsaw Puzzle," for example, requires a meaningful pattern at finally by placing certain pieces in settled spatial position and orientation using robot cell. We can summarize that "Jigsaw Puzzle" focuses on spatial reasoning rather than temporal reasoning sine chronological operations to finish the puzzle are needless during the whole placing process. "Tic-tac-toe Game" is the contrary that emphasizes moving chess chronologically rather than its spatial position and orientation (distinguish the type of pieces rather than each piece individual). Claw machine is another popular game that involves picking and placing to clear the toy tray. We hypothesize that both robot cells and intelligent algorithms lead to performance differences when executing game manipulation tasks.

\subsection{Functional Pipeline}
\begin{figure}[htbp]
    \centering
    \includegraphics[width=1\columnwidth]{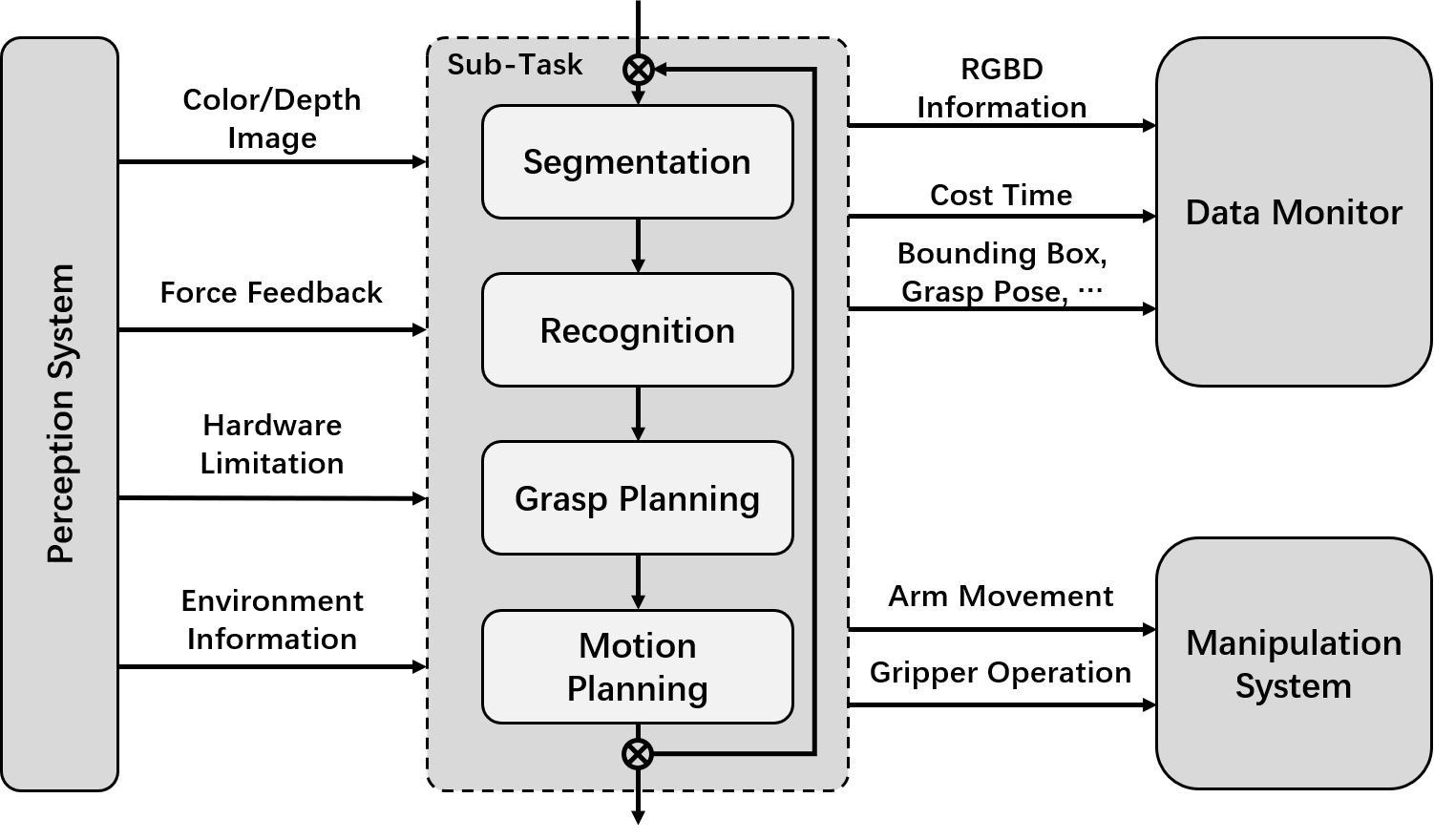}
    \caption{Functional pipeline of the DeepClaw for structured data-driven benchmarking, consists of four steps: segmentation, recognition, grasp planning and motion planning. Sensors, which are parts of perception system, provide plentiful perceptual information to algorithms among steps. Data monitor collects valuable data during the process. Actions are given to the manipulation system.}
    \label{fig:FunctionalPipeline}
\end{figure}

In DeepClaw, a sub-task is defined by a pipeline of modules, including segmentation, recognition, grasp planning, and motion planning, as shown in Fig \ref{fig:FunctionalPipeline}. The pipeline takes color/depth images, force feedback, hardware limitation, and environment information as input and gives actions to the manipulation system and pushes data and results to data monitor. 

Segmentation and recognition involve analyzing information gained from the perception system. Segmentation is the process that robot cell collecting environment information and representing spatial information of the target objects by using perception algorithms. The output of the segmentation module can be pixel-wise masks or bounding boxes. DeepClaw includes basic segmentation based on contour and edge detection in Opencv \cite{opencv_library}. 

Recognition is the process of extracting features of the target object beyond location information. In this step, the robot cell infers the category of the target object by applying specific methods, such as support vector machine(SVM) and convolutional neural network \cite{krizhevsky2012imagenet}. Some of the end-to-end neural networks infer the location and category of the target object at the same time \cite{liu2016ssd, he2017mask}.

Grasping planning aims to find the optimal pose for the robot arm and end-effect to approach the target objects, which is highly dependent on both the end-effector and the objects. Recent years, research interests have shifted from analytic method \cite{miller2003automatic,pokorny2013classical} to data-driven method \cite{mahler2017dex, pinto2016supersizing,kappler2015leveraging}. DeepClaw has implemented an end-to-end grasp planning model based on fully convolutional AlexNet, which was trained on 5000 random grasps with labels.

Motion planning utilizes information above, such as grasping pose, force sensor data, constrain of the robot system, and limitation of working space, to obtain collision-free trajectories. Currently, waypoint-based motion planning is used through our tasks. For UR5 and UR10e, we utilize the movej command implemented in UR's controller to plan and execute a path between waypoints. For Franka, we utilize a fourth-order motion generator in the joint space provided by the libfranka software.

Researchers can assess different robot cells in the same manipulation task by standardizing sub-tasks in the above DeepClaw pipeline. Moreover, a new algorithm for each functionality of the pipeline can be evaluated by implementing the algorithm as a module and integrate it into the task pipeline. Besides, DeepClaw pipeline provides the possibility to analyze the variance between end-to-end neural network models and traditional step-by-step algorithms.

\subsection{Result Reporting}
One of the key problems in benchmarking is consistent evaluation metric and automatic data collection process. The DeepClaw provides a monitor module to record the experiment data automatically and keep a snapshot of the necessary codes and configurations to be able to reproduce and reference the experiments in the future. Evaluation metrics can be divide into function-wise and task-wise. The former metric only evaluates a single functionality of the pipeline. Examples are Intersection-over-union (IoU) for segmentation \cite{he2017mask}, recall and precision for recognition \cite{powers2011evaluation}, force-closure for grasp planning \cite{nguyen1988constructing}, length and clearance for motion planning \cite{sucan2012open}. These metrics require a good knowledge of the truth hence require extensive manual annotations. The latter is defined to evaluate the completeness and efficiency of the task as a whole. A universal and easy-to-get performance metric is the time cost of each step in the task pipeline, which is a key metric when comparing between different algorithms and robot platforms. DeepClaw has implemented the time cost metric, task-wise metrics such as success rate of grasping, and some function-wise metrics, which will be described in Section IV.

The raw data produced by the robot arm and sensors, e.g., robot poses, RGB images and depth images, is also recorded throughout the pipeline depending on the tasks requirement. The raw data and metric data are all in well defined fommat, hence the same analysis script can be used to generate

\section{Experiment Benchmarks}
\label{sec:ExpBenchmark}

We conduct three benchmarking tasks as examples of applications based on DeepClaw. Each task represents a specific task of board games, assembly, and grasping, respectively, and is repeated on all the three robot cells we have built. 

\subsection{Tic-Tac-Toe: Board Games as Adversarial Interaction}
\begin{figure}[htbp]
    \centering
    \includegraphics[width=1\columnwidth]{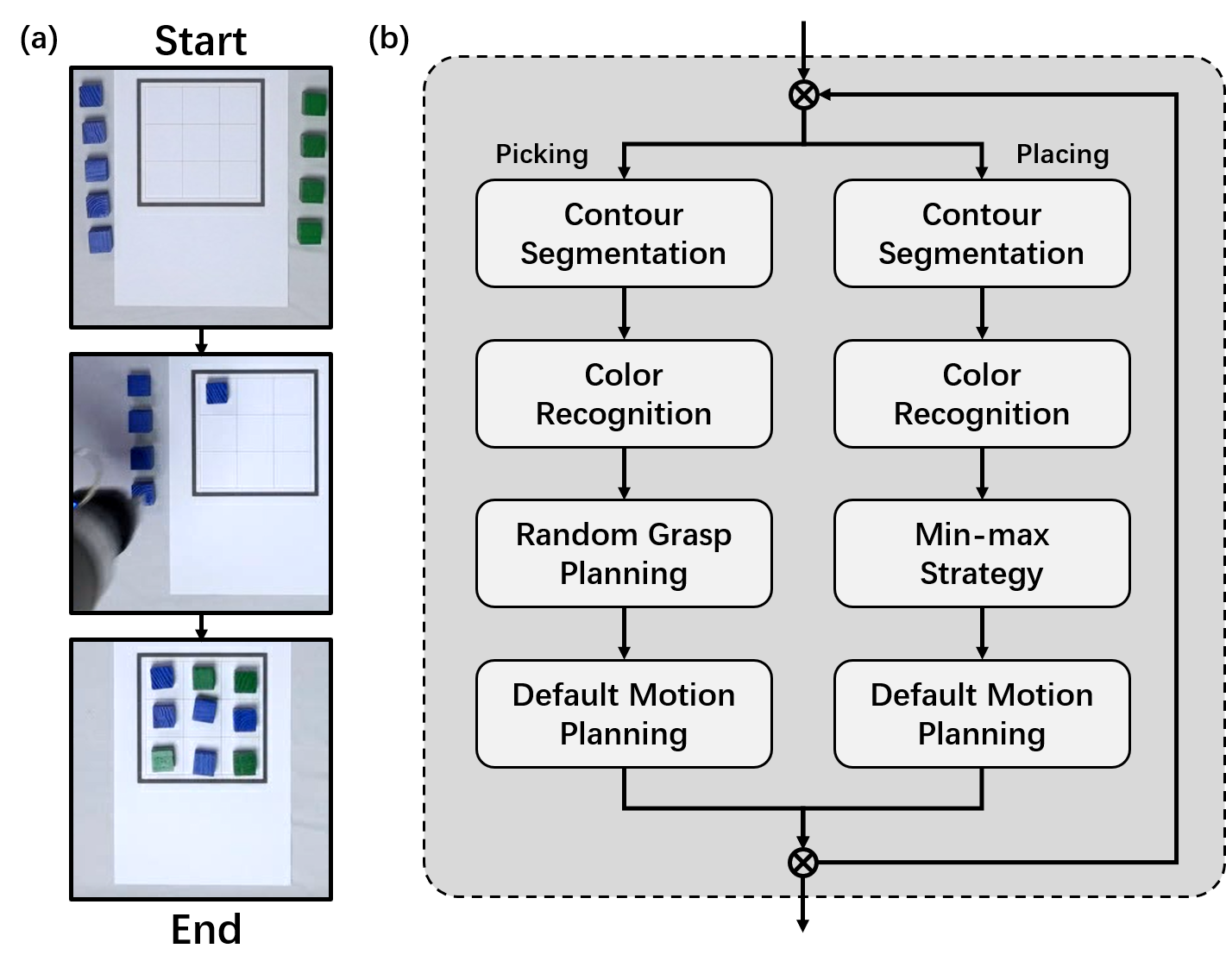}
    \caption{(a) The first row shows initial configuration and the third row shows the state of Tic-Tac-Toe task completion. (b) The pipeline of the baseline Tic-Tac-Toe benchmark.}
    \label{fig:ExpBaseline-Design}
\end{figure}

In this section, we elaborate on how we implemented the Tic-Tac-Toe game as a benchmark test for adversarial interaction. Tic-Tac-Toe game is a temporal reasoning related task, which required two players moving pieces alternately. To simplify this game as a baseline, the two players use the same placing strategy, namely the Minimax algorithm with depth 3, and are both executed by the robot arm. We use green and blue cubes from Yale-CMU-Berkeley objects set \cite{calli2015benchmarking} representing two types of pieces. At the start of the game, 3$\times$3 checkerboards printed on an A4 paper is placed in front of the robot base, and the two types of pieces are lined on the left and right side of the chessboard as shown in Fig \ref{fig:ExpBaseline-Design}(a). The task is to pick a type of piece and place it on one of nine boxes on the checkerboard in turns until one player wins or ends with a tie. 

The full task of the Tic-Tac-Toe game can be divided into two repetitive sub-tasks, as shown in Fig \ref{fig:ExpBaseline-Design}. In the picking sub-task, we locate the cubes from contour segmentation and select one the cubes with the same color randomly. In the placing sub-task, we use the Minimax algorithm to determine the placing location. The time to complete a pick and place sub-task $t_{sub}$ is recorded as a performance metric. To compare the performance of the robot arm, we purposely exclude the gripper closing time from $t_{sub}$. We performed the bin-clearing task ten times on each robot cell, and the performance metric is averaged over ten repeated tasks.

The results show that all the repeated tasks were completed, given the simplicity of the game and the robustness of the contour segmentation. The time costs of a pick and place sub-task are 9.6 s, 17.9 s, and 19.5 s for Franka, UR5, and UR10e, respectively. Note in this benchmark, the joint speed and acceleration limits are set to 0.7 $m/s$ and 1.6 $m/s^2$ for UR5 and UR10e and are set to 1 $m/s$ and 2.5 $m/s^2$ for Franka. With the same speed setup, UR5 is slightly more efficient than UR10e. Franka cut down almost half of the time with 1.5 times joint speed.

\subsection{Claw Machine: End-to-End Manipulation Benchmarking}
\begin{figure}[htbp]
    \centering
    \includegraphics[width=1\columnwidth]{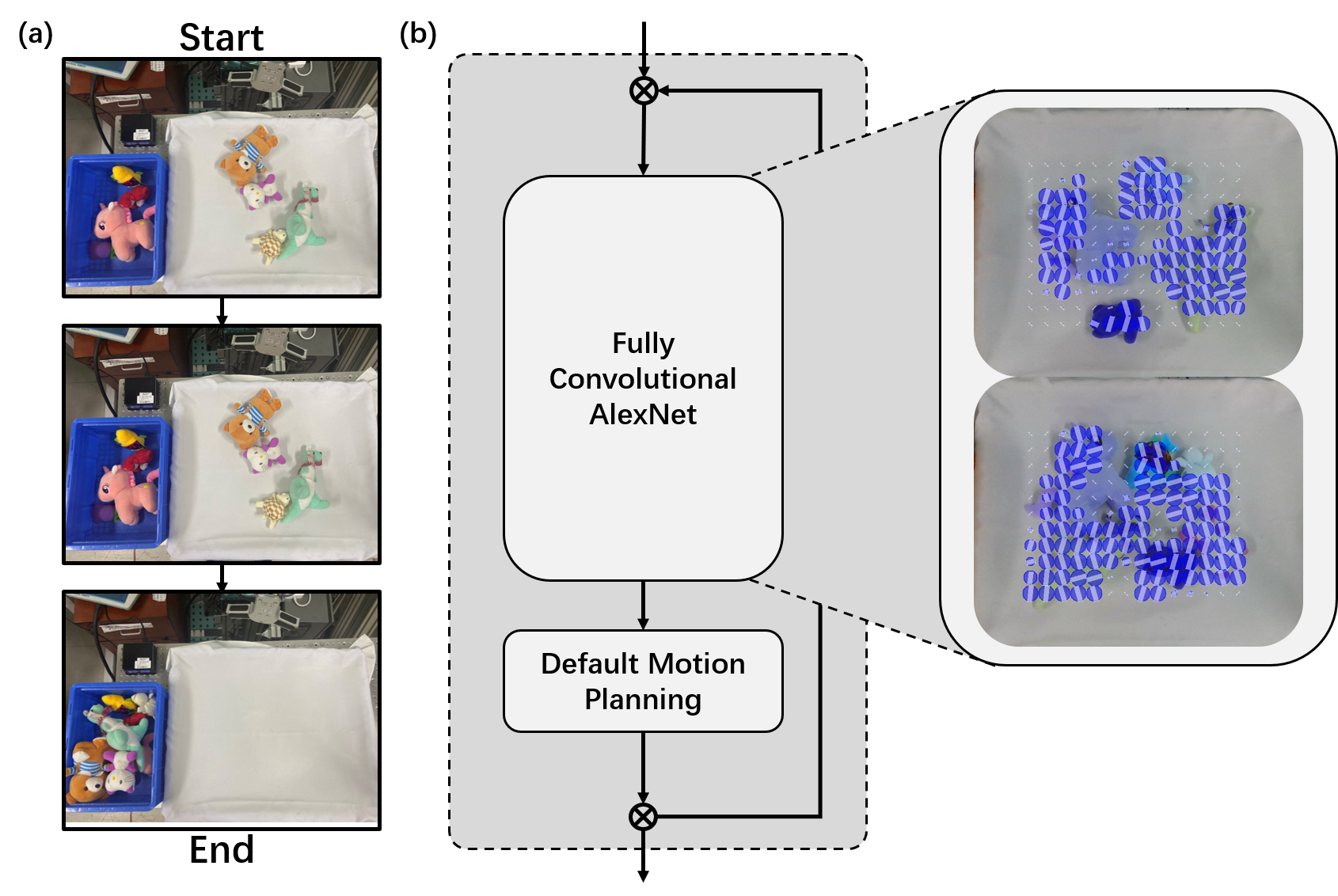}
    \caption{(a) The first row shows initial configuration and the third row shows the state of claw machine task completion. (b) The pipeline of claw machine benchmark.}
    \label{fig:ExpClaw-Design}
\end{figure}

In this section, we used a claw machine scenario to illustrate a bin-clearing task. This benchmark measures the performance of a learned policy for predicting robust grasps over different robot cells. At the start of the task, a 60cm$\times$70cm white bin stuffed by eight soft toys and an empty 30cm$\times$40cm blue bin are placed side by side on the table top as shown in Fig \ref{fig:ExpClaw-Design}. The task is to transport the toys to the blue bin one by one until clearing the white bin. We restrict the gripper to grasp vertically, allowing only rotations along the z-axis of the robot base. 

Each sub-task contains a pick and place cycle. The pipeline predicts the optimal grasping pose using the fully convolutional AlexNet trained from a dataset of 5000 random grasps of soft toys. We replace the last three fully connected layers by $1\times1$ convolutional layers so that the model can process images with various resolutions. The CNN model takes an RGB image as input and outputs relatively dense map of success probabilities of grasps at pixel level. At each pixel with prediction, the model give success probabilities for 18 rotation angular bins, the angle with the highest success probability is selected, as shown in Fig. \ref{fig:ExpClaw-Design}(b), where the radius of circles represents the probability, and the white lines indicate grasp orientations. The optimal grasp center $(u,v)$ and the associated orientation $\theta$ is then obtained by finding the pixel with the highest probability in the image space. Then grasp center $(u,v)$ is transformed to $(x,y)$ with reference to the robot base using depth at $(u,v)$ from the depth map, the intrinsic parameters of the camera, and hand-eye matrix. The grasp depth $z$ is fixed slightly above the table top. For performance assessment, we report robot arm execution time $t_{pick}$ for a single pick task, and the total picking success rate $r_{success}$, which equals to eight over the number of total grasp attempts before clearing the bin. We performed the bin-clearing task on three robot cells and ten times on each robot cell. The performance metric is average over ten repeated tasks. We also excluded the gripper closing time from $t_{pick}$ to compare the efficiency of robot arms. The closing time is 0.80 s, 2.0 s, and 0.33 s for Franka, RG6, and HandE, respectively.

The picking success rates for Franka, UR5, and UR10e are 0.80, 0.91, and 0.83, respectively. The results demonstrate the grasp abilities variation of Franka, RG6, and HandE, among which RG6 is most capable due to its sizeable graspable area. The robot picking times $t_{pick}$ are 8.3 s, 8.6 s and 11.2 s respectively for Franka, UR5 and UR10e. In this benchmark, the joint speed and acceleration limits are set to 0.7 $m/s$ and 1.6 $m/s^2$ for all three robot arms. Franka and UR5 perform similarly while UR10e takes the longest time as in the Tic-Tac-Toe benchmark.

\subsection{Jigsaw Puzzle: Tiling Game for Modular Benchmarking}
\begin{figure}[htbp]
    \centering
    \includegraphics[width=0.8\columnwidth]{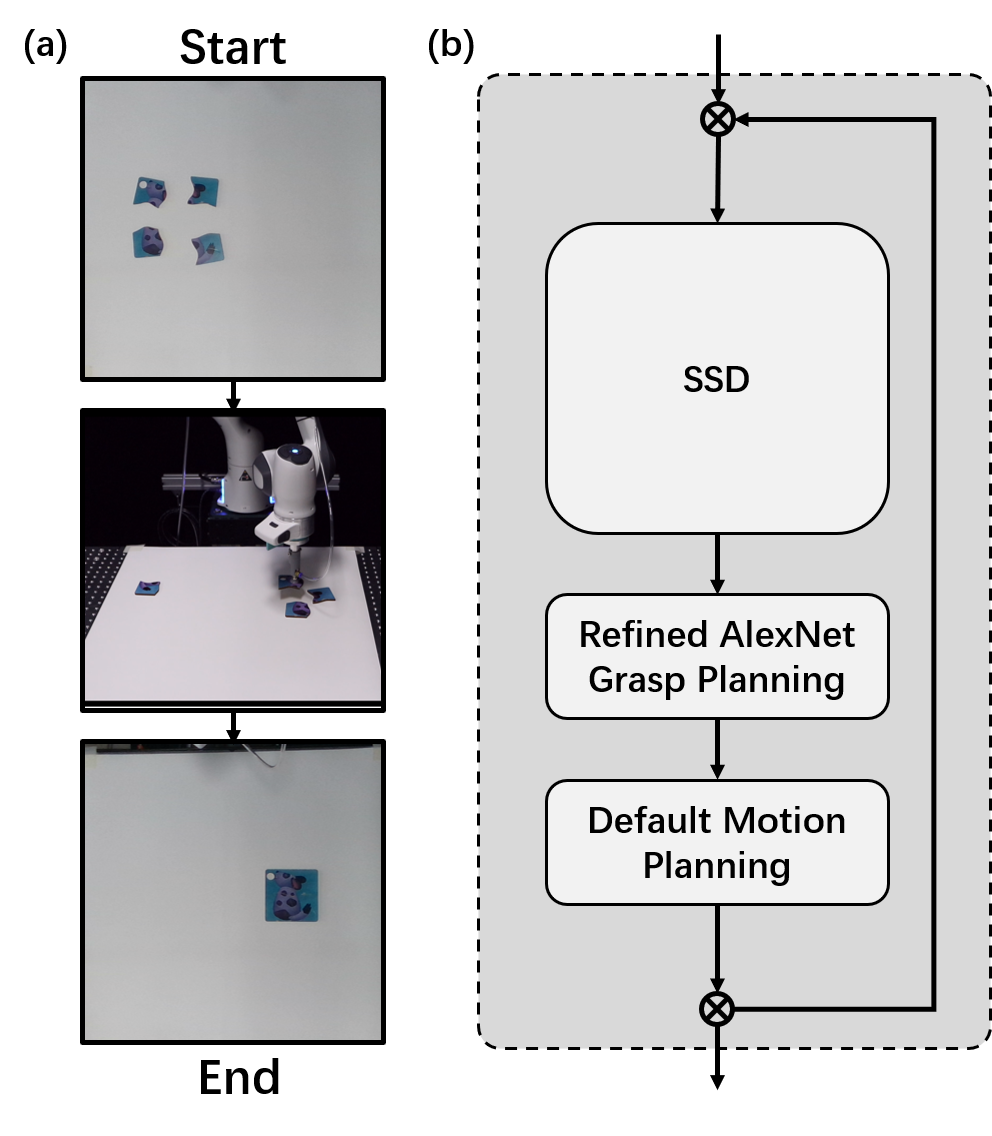}
    \caption{(a) The first row shows initial configuration and the third row shows the state of jigsaw task completion. (b) The pipeline of the jigsaw benchmark.}
    \label{fig:ExpJigsaw-Design}
\end{figure}

The jigsaw puzzle benchmark is designed to evaluate learning models for object detection and recognition for tiling tasks. A jigsaw puzzle is a tiling game that requires the assembly of often oddly shaped interlocking  and tessellating pieces. The jigsaw set used in this paper contains four thin wooden pieces with an image printed on one side and can form a 10.2cm$\times$10.2cm picture when they are correctly assembled. We use a suction cup to complete the task on all three robot cells as the jigsaw piece is only 5 mm thick and is too challenging for grippers. At the start of the task, the four pieces are randomly placed on the table top, as shown in Fig. \ref{fig:ExpJigsaw-Design}(a). The task is to detect and pick one jigsaw piece at a time and place it at the required location according to its shape and texture information, and finally assemble all the four pieces into one whole piece. We restrict the gripper to pick vertically, allowing only rotations along the z-axis of the robot base. 

The pipeline uses a fine-tuned Single-Shot Detection (SSD) model \cite{liu2016ssd} for segmentation and recognition. SSD gives a bounding box of the piece without orientation and determines which one of four pieces it belongs. Also, a fine-tuned AlexNet is adopted to predict the grasping rotation angles. We collected about 80 images for the training of SSD and used data augmentation to avoid overfitting. The training data for AlexNet was automatically generated by rotating four template images. For performance assessment of the overall task, we report a 2D measurement of the area rate defined by

\begin{equation}
    score = \frac{A_{standard}}{A_{actual}}
    \label{eq:EyeOnBase}
\end{equation}
to evaluate the task completion, where $A_{standard}$ refers to the area of the finished jigsaw when the four fragmented pieces are perfectly aligned together, $A_{actual}$ refers to the actual area of the minimum bounding box of the finished four jigsaw pieces. For performance assessment of functionalities in the pipeline, we report the IoU for segmentation, average precision (AP) for recognition, the success rate of picking and time cost of a pick and place sub-task as shown in Table \ref{tab:ExpJigsaw-Results}. 

The results compare the performance of the task on the three robot cells. The IoU and AP are comparable on UR5 and UR10e and drops on Franka, possibly because the training data of the SSD model are collected on D435. The result is expected to improve on Franka if we include images from D435i as well. In this benchmark, the joint speed and acceleration limits are set to the same for the three cells. However, the execution time of Franka is significantly less than UR5 and UR10e. A plausible explanation is that 7 DoF robot arm is more agile and faster than 6 DoF robot arms for more complex manipulation tasks.

\begin{table}[htbp]
    \caption{The jigsaw benchmark results for modularized tiling tasks.}
    \label{tab:ExpJigsaw-Results}
    \begin{center}
        \includegraphics[width=1\columnwidth]{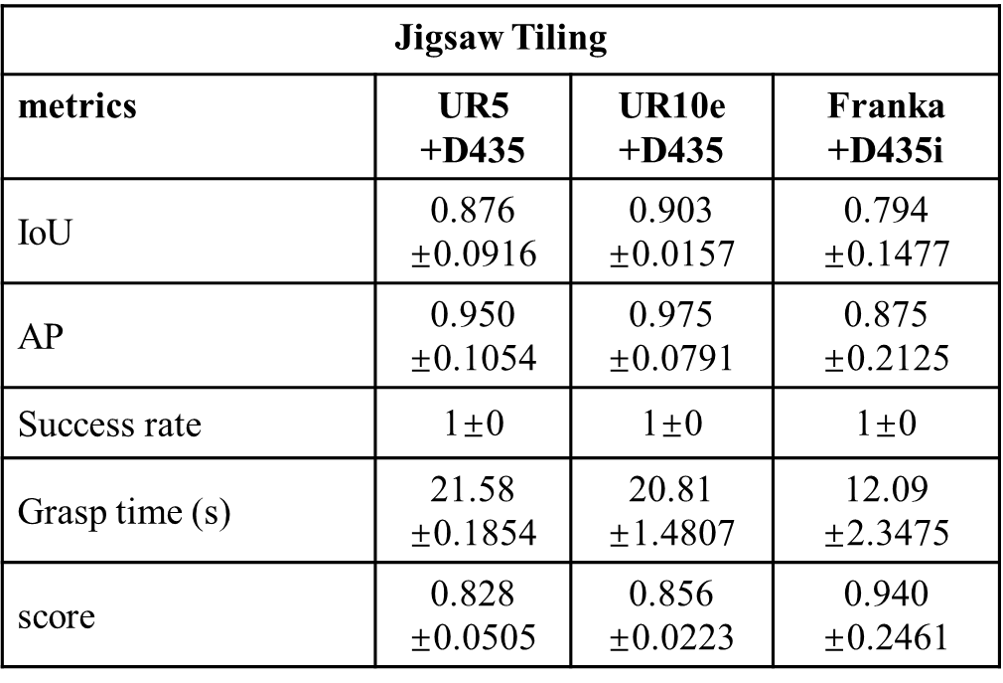}
    \end{center}
\end{table}

\section{Conclusion and future work}
\label{sec:Conclusion}
This paper proposes DeepClaw, a minimum robot cell design, and software to accelerate benchmarking the robotic hardware and learning manipulation. To illustrate the use of such hardware and software design, we have integrated three robot cells involving three different robot arms, four cameras, and four end-effectors. We demonstrate three benchmarking tasks and our experiments show that tasks can be reproduced easily with different hardware by standardizing the pipeline of functionalities and the result reporting. However, the performance of the task depends on the hardware.

While our effort is to provide a standard design of hardware and algorithm pipeline, DeepClaw currently only features three benchmarking tasks on three different robot cells. This is a significant limitation whenever deals with hardware and can only be resolved by encouraging collaborations within the community. Future work will be focused on three directions: 1) adding modules of real-time robotic control with close-loop such as visual servoing \cite{Janabi2011Comparison}, force control \cite{Liu2004Real}, and motion planning and 2) incorporating existing simulation effort from the community like RLBench \cite{james2019rlbench} and 3) expanding the benchmarking tasks and dataset. There are several ongoing projects based on DeepClaw in our lab, including benchmarking new gripper designs for manipulation and visual–tactile sensor fusion, and we shall continuously update DeepClaw with our new results. We hope DeepClaw will contribute to and invite collaborations from the robotics and learning communities.

\bibliographystyle{unsrt}  
\bibliography{references}  

\end{document}